# LLM vs. Lawyers: Identifying a Subset of Summary Judgments in a Large UK Case Law Dataset

Ahmed Izzidien,[*][†] Holli Sargeant,[*] and Felix Steffek[*]

## Abstract

To undertake computational research of the law, efficiently identifying datasets of court decisions that relate to a specific legal issue is a crucial yet challenging endeavour. This study addresses the gap in the literature working with large legal corpora about how to isolate cases, in our case summary judgments, from a large corpus of UK court decisions. We introduce a comparative analysis of two computational methods: (1) a traditional natural language processing-based approach leveraging expert-generated keywords and logical operators and (2) an innovative application of the Claude 2 large language model to classify cases based on content-specific prompts. We use the Cambridge Law Corpus of 356,011 UK court decisions and determine that the large language model achieves a weighted F1 score of 0.94 versus 0.78 for keywords. Despite iterative refinement, the search logic based on keywords fails to capture nuances in legal language. We identify and extract 3,102 summary judgment cases, enabling us to map their distribution across various UK courts over a temporal span. The paper marks a pioneering step in employing advanced natural language processing to tackle core legal research tasks, demonstrating how these technologies can bridge systemic gaps and enhance the accessibility of legal information. We share the extracted dataset metrics to support further research on summary judgments.

**Keywords**: large language models, computational legal methods, legal corpora, UK case law, LegalAI, summary judgment, regular expression

[*] Faculty of Law, University of Cambridge; Cambridge, United Kingdom.
[†] ai297@cam.ac.uk

This work was funded by the Nuffield Foundation grant Access to Justice Through Artificial Intelligence. The views expressed are those of the authors and not necessarily of the Foundation. Visit www.nuffieldfoundation.org. We are grateful to Nicola Mathew for excellent research assistance.

# 1 Introduction

Finding the perfect precedent amidst the extensive volume of case law is a challenging yet crucial enterprise. For practitioners and academics alike, this process is not unlike seeking a needle in a haystack – meticulous and time-intensive (Stefancic 1997; Kleinman 2019). It requires sifting through countless cases to uncover the one judgment that could decisively inform a legal argument or shed light on a scholarly inquiry.

Legal research is especially complicated when attempting to identify all cases that deal with a certain legal issue (Olsen and Esmark 2020). We identified this challenge in a broader line of research on a type of legal proceeding called summary judgment in the United Kingdom (**UK**). In the UK, cases are not automatically categorised by legal issues. Searching case text for the term "summary judgment" is not sufficient to reliably identify true summary judgment cases because judgments frequently reference other related cases or areas of law, and discuss the prior history of a case, which might include a prior application for summary judgment. The multifaceted nature of these cases, in which multiple legal issues often intertwine, compounds the difficulty of systematically extracting subsets of a large legal corpus for further analysis.

We focus on summary judgments due to their critical role in allowing courts to resolve cases without a full trial if there is "no real prospect" of success at trial and there is no other "compelling reason" for such a trial. Summary judgments are an important procedural device in civil litigation, empowering courts to resolve disputes efficiently. However, summary judgment applications are disproportionately brought against self-represented litigants who often lack the legal expertise to oppose them successfully (Leitch 2018). These considerations have prompted some scholars to call for abolishing the procedure (Bronsteen 2007; see also Gelbach 2014). Empirical studies in Canada confirm the disparity, finding near-perfect summary judgment grant rates against self-represented litigants compared to 60-70% when both parties have legal representation (Macfarlane et al. 2015). Growing numbers of summary judgments raise considerable challenges. The balance between simplified procedures and proportionate judgments must be balanced with a commitment to access to justice.[1]

While these concerns have arisen from research in jurisdictions other than the UK, little research exists to assuage fears that the trend is being replicated in the UK. To better understand UK trends in summary judgment practice, we first need to reliably identify these decisions within the vast body of case law in the UK.

Our study utilises the Cambridge Law Corpus (**CLC**) that aggregates UK court judgments from 1595 to 2023 in its latest version (Östling et al. 2023). As such, this paper aims to accurately identify all summary judgment cases in the CLC for future research and contribute to the methodology of identifying subsets of case law in large legal corpora.

Against this backdrop, our study addresses two foundational research questions:

---

[1] *Hyrniak v Mauldin* [2014] 1 SCR 87, at 93 and 99 (Canada).

1. How can a specific subset of cases, such as summary judgments, be identified within a large legal corpus of mixed cases?

2. How are summary judgment cases distributed across time and different UK courts?

To answer the first question, the paper presents two computational methodologies for extracting case subsets. Initially, we engage legal experts to analyse the language patterns characteristic of summary judgment cases using traditional natural language processing (**NLP**) methods. They construct a search term matrix using logical operators. Our second method employs the large language model (**LLM**) Claude 2. This model reviews uploaded cases and, through a bespoke prompt, determines their classification as a summary judgment or not. We compare both methods and assess the effectiveness of their ability to identify and isolate summary judgment cases (Anthropic 2023a).

To answer the second question, once we have isolated the subset of summary judgment cases, we extract the names of the courts and hearing dates from each case. We use visual representations of our results.

Our contributions are twofold. First, we offer a comparative analysis of two computational methods to isolate summary judgment cases from a large legal corpus. Second, we provide a depiction of how summary judgment cases are distributed in terms of time and court hierarchy.

This investigation marks a significant stride in applying computational techniques to a fundamental legal research task: identifying subsets of legal cases within large corpora. Thoughtful application of these technologies can diminish systematic obstacles and democratise access to legal resources.

## 2    Related Work

Several legal corpora containing court decisions have recently been released to support NLP research on legal texts. For our purposes, we use the recently published CLC, which includes judgments from the UK (Östling et al. 2023). In other jurisdictions, MultiLegalPile by Niklaus et al. (2023b) includes US, European and Canadian court decisions, and LEXTREME by the same group collates 11 datasets comprising over 100 million legal documents (Niklaus et al. 2023a). Chalkidis et al. (2023) introduced LexFiles, a legal corpus covering legislation and case law from six English-speaking jurisdictions, including the EU, US, Canada and India. In terms of contract corpora, the CUAD dataset by Hendrycks et al. (2021) contains over 13,000 annotated legal contracts. To support language model pretraining on legal texts, Chalkidis et al. (2020) released legal-BERT trained on publicly available legal corpora, while Zheng et al. (2021) analysed a 37GB corpus of Harvard Law cases for further pretraining. In terms of non-English legal corpora, Hwang et al. (2022) introduced a Korean court judgment corpus, and Poudyal et al. (2020) previously released a corpus of European Court of Human Rights judgments.

The introduction of these legal corpora allows new computational methods to evaluate several aspects of the law. There is a wealth of literature related to NLP and law. Many focus on techniques for legal text classification, sentiment analysis and topic modelling, as well as the use of machine learning (**ML**) to predict judicial decisions and analyse legal texts (Hendrycks et al. 2021; Medvedeva et al. 2023).

There have also been ongoing attempts since the 1990s to extract information from legal texts (Turtle 1995). Information extraction is a method to "extract instances of predefined categories from unstructured data, building a structured and unambiguous representation of the entities and the relations between them" (Nadeau and Sekine 2007). It has proved to be a challenging method for working with unstructured and Big Data (Adnan and Akbar 2019).

One particular challenge this paper aims to address, is to evaluate whether computational methods can be applied to find relevant cases within a corpus or entire subsets of case law. To the best of our knowledge, there is no other literature on identifying entire subsets of case law from within a large legal corpus.

We have identified the different, but related task of case law retrieval, whereby computational methods are used to identify supporting cases. An annual competition on legal information extraction and entailment (**COLIEE**) (Goebel et al. 2023) has highlighted the difficulties researchers face when attempting this task. The most recent results from this competition reflect the associated difficulties, with F1 scores ranging from 0.17 to 0.38 out of 1 (Li et al. 2023). The F1 score is a metric that combines precision and recall into a single measure of a model's performance (Christen et al. 2023). It provides a harmonic mean of precision and recall, effectively balancing these two metrics. The term "precision" refers to the proportion of predicted positive cases that are correctly identified. It measures how many cases the model predicts as positive are true positives. High precision means few false positives.

On the other hand, the term "recall" refers to the proportion of true positive cases that are correctly predicted. It measures how many of the total positive cases the model identifies. High recall means few false negatives. Precision and recall often trade off against the other because improving one may degrade the other. Precision could be made perfect by predicting only obviously positive cases, but recall would likely suffer. Recall could be perfected by predicting all cases as positive, but precision would be poor. As such, the F1 score accounts for both metrics by taking their harmonic mean. If either precision or recall is low, the F1 will be low. It is maximised only when both measures are high.

In the COLIEE competition, a higher score was achieved using a structure-aware pre-trained language model. Such models use heuristic pre- and post-processing approaches to reduce the influence of irrelevant items. They also use learning-to-rank methods, a ML method employed to rank documents according to their relevance (Li et al. 2023).

A recent paper on prior case retrieval, a process of automatically citing relevant prior legal cases in a given query case based on facts and precedent, achieved an F1 score of 0.39 out of 1. The researchers achieved this by extracting events from documents and using events for retrieval, where events were defined as predicate-argument tuples capturing key actions (Joshi et al. 2023).

These results highlight the persisting difficulties faced in large-scale legal case retrieval.

# 3      Data

The data used for this study consists of UK case law from the CLC (Östling et al. 2023).[2] The CLC is a corpus containing court decisions, also called judgments or cases, from the combined UK jurisdiction, including England and Wales. The UK has three distinct jurisdictions: England and Wales, Scotland and Northern Ireland. Some courts or tribunals hear cases from across the three jurisdictions, which fall into the combined UK jurisdiction.

The CLC consists of 356,011 cases from UK courts and tribunals, spanning from 1595 until June 2023 (in its latest version). However, the corpus does not contain all UK cases. Shubber, Hoadley and others have observed that 'black holes of information are common in English law' (Shubber 2022; Hoadley et al. 2022; House of Commons Justice Committee 2022). Courts have discretion as to what judgments are published. For example, a study found that between 2015 and 2020 only 55% of 5,408 administrative court judgments identified by a commercial subscription service appeared on the website of the British and Irish Legal Information Institute (**BAILII**) (Somers-Joce et al. 2022; Byrom 2023). This could be a barrier to the data-driven analysis of English law (Janeček 2023).

However, analysis of available law is important. Especially as the common law of England and Wales is one of the major global legal traditions and has been adopted or adapted in numerous countries worldwide. Common law jurisdictions mean that authoritative law includes both legislation and court judgments. Legislation comprises Acts of Parliament (under the authority of the Crown) and regulations. Judges apply legislation and case law principles decided in previous cases to explain their decisions. Judges in different courts may decide cases with different principles across these three jurisdictions. Court judgments may be appealed to the highest court – the UK Supreme Court (formerly the House of Lords).

In the common law tradition, court decisions can operate as an independent primary source of substantive law (case law), meaning that courts can legally justify their decisions by applying case law in the same way they apply legislation (Raz 2009). Lawyers often call past cases *precedents*, and in many contexts, precedents at least influence the decisions of courts when relevantly similar disputes arise. In Anglo-American legal systems, this legal principle is known as *stare decisis* (Duxbury 2008, p. 12). Courts in the UK apply *stare decisis* relatively strictly.

The CLC provides a valuable resource for studying English case law. Each case in the CLC is held in extensible markup language (XML), including a set of tags known as namespaces to describe the text. The tags include the court name, the last date of the hearing's delivery, the neutral citation and the case text.

# 4      Methodology

We compare two methodologies to achieve our goal of identifying and isolating all summary judgment cases within the CLC. First, we use domain knowledge to identify summary judgment cases. We implement this by using the

---

search logic we develop after the initial exploration of the data. The method is well-tested and is often used to augment other processes in information retrieval (Pudasaini et al. 2022; Kanhaiya et al. 2023).

Second, we propose an innovative method combining legal expertise with advanced NLP technologies. We use Claude 2, an LLM that can analyse long and complex texts (Anthropic 2023a). We craft a detailed prompt outlining the key attributes of a summary judgment case. Claude 2 then assesses each relevant case in the corpus, classifying it as either containing summary judgment proceedings or not based on explanatory factors. This innovative use of legal knowledge to harness Claude's reasoning presents a novel technique for legal case identification.

### 4.1    Using Expert Domain Knowledge

### 4.1.1    Legal context of summary judgments

Summary judgments aim to resolve a persistent challenge of common law litigation that often results in considerable delays and large expenses. While summary judgment did not exist in common law, it evolved through a series of statutory interventions that were intended to respond to social and economic pressures impacting the free flow of commerce (Clark and Samenow 1929; Bauman 1956; Halligan 2022). The evolution of the procedure focused on factual clarity and evidence; summary judgment was only granted to plaintiffs in cases approaching factual certainty (Haramati 2010). Early procedures have developed since the 1800s,[3] which have evolved into the modern procedures contained in the current *Civil Procedure Rules 1998* (**CPR**).[4] Therefore, we only aim to identify cases based on the most current procedure and only search for cases from 26 April 1999 onwards. Even if we had not limited our search, it is unlikely that we would have identified summary judgments with the same techniques as the language of the law has evolved considerably over time. In particular, the 1954 procedure was called "trial without pleadings" instead of "summary judgement".

Identifying relevant case law is a core skill lawyers are required to possess and is unique to the profession making it more challenging than search approaches to other datasets (Hutchinson 2008; Valentine 2009). First, the objective is typically to locate seminal cases that establish legal doctrines and tests rather than identify an exhaustive list of potentially applicable cases. Locating key cases requires identifying a specific legal issue rather than casting a wide net, an atypical approach in academic research that normally prioritises comprehensiveness. Second, legal language demonstrates considerable variability across different practice areas and the individual writing styles of judges. An effective researcher must adapt search terminology to the relevant legal context and individual judicial preferences.

Law is a field that relies heavily on the use of notoriously complex, domain-specific language (Ruhl 2008; Katz and Bommarito 2014; Friedrich 2021). Legal language tends to be more specialised than other kinds of natural language. Legal terms and idioms, which are often difficult for a layperson to understand, have strong semantics and are often the keystone of legal reasoning (Dale 2017; Nazarenko and Wyner 2017). The complexity of legal language and massive volumes of textual data have made it difficult to develop NLP systems that understand the nuances of legal tasks (Katz et al. 2020; Chalkidis et al. 2022).

---

[3] Summary Procedure on Bills of Exchange Act 1855 (UK) (18 & 19 Vict c 67) [Keating's Act]; Business of Courts Committee *Interim Report* (Cmd 4265, 1933) at [15]-[17]; Annual Practice 1956, Order XIV B is discussed in 99 SOL. J. 157-58 (1955).
[4] Civil Procedure Rules 1998, pt 24 (in force 26 April 1999); made under authority of the Civil Procedure Act 1997.

To address this complexity, we began with an exploratory data analysis to build the summary judgment dataset.

### 4.1.2    Exploratory analysis of the dataset

After obtaining the CLC corpus, we loaded it into a Python DataFrame. We used the XML tags present in the corpus to save each case with its date, neutral citation number and court name in a Pandas DataFrame.

First, we aimed to identify any case in the CLC corpus that referred to a summary judgment. We used a "regular expression" (RegEx) search. RegEx is a powerful language used for pattern matching and searching within text (Jurafsky 2014). Formally, it is an algebraic notation for defining specific sequences of characters, essentially capturing a set of strings. This tool is particularly useful for text analysis, especially when one needs to locate instances of a pattern across a large body of text, such as a corpus. A RegEx search function will search through an entire corpus and identify all occurrences of the pattern in question (Jurafsky 2014).

We used this method to search for the term "summary judgment" in the CLC. To account for potential variations in spelling or format, the RegEx pattern implemented was: **"r'\bsumm[a-z]*\s*judg[a-z]*'"**. This expression is designed to find the root words "summ" and "judg" followed by any number of alphabetical characters, accounting for possible differences in the way the term is written (i.e., "summary judgment", "summary judgement", "summary judgments" etc.). We also set the search to be case-insensitive. The approach is intended to strike a balance between specificity and flexibility. The results were saved as the RegEx SJ dataset.

While an important first step, cases that contain the words "summary judgment" will not inherently be actual summary judgment cases. One reason for this is that judgments will often refer in-text to other judgments, either by different courts or by different judges in the same matter. For example, a case on appeal may mention that at an earlier stage, one of the parties had applied for summary judgment, but that is not the current issue before the court. To identify a true summary judgment, lawyers would use more detailed aspects of legal research, such as using multiple keyword searches and referring to the relevant legislation. The legal experts, co-authors of this paper, identified a series of relevant keywords for the below search matrix (Table 1).

### 4.1.3    Search matrix to identify summary judgments

The search matrix was developed by legal experts by reference to the CPR, seminal case law on summary judgment and practice notes on summary judgment (Lexis+ 2023; Thomson Reuters Practical Law 2023). After conducting legal research on the important elements of bringing a summary judgment application and what factors a court considers in making its determination, relevant terms and phrases were identified that should be present in summary judgment cases. These we called "keywords". The keywords are ordered by category, where each category lists several variations of keyword terms. For example, the category "summary judgment" includes the variations "summary judgement" and "summary judgment".

A second RegEx search using these terms was conducted on the RegEx SJ dataset. The result of this new search was used to build a count and co-occurrence matrix. The matrix captures the co-occurrence of the keywords given in Table 1. Such a matrix shows how often the keywords listed occur together in a set of documents.

*Table 1: Keyword search terms based on terms used in summary judgments*

| Category | Keyword search terms |
|---|---|
| **summary judgment** | summary judgement |
| | summary judgment |
| **compelling reason** | compelling reason why the case or issue should be disposed of at a trial |
| | compelling reason to try the case or issue |
| **civil procedure rules part 24** | civil procedure rules part 24 |
| | cpr 24 |
| | cpr 24.2 |
| | cpr part 24 |
| | part 24 of the civil procedure rules |
| | part 24 application |
| | part 24 judgement |
| | part 24 judgment |
| | r 24.2 |
| | r. 24.2 |
| | rule 24.2 |
| **easyair v opal** | easyair v opal |
| | easyair ltd v opal telecom |
| | easyair ltd. (t.a openair) v. opal telecom ltd [2009] ewhc 339 (ch) |
| | ewhc 339 (ch) |
| **real prospect of success** | real prospect of success |
| | real prospect of succeeding |
| | realistic prospect of success |
| | realistic prospect of succeeding |
| | no real prospect |
| | no real prospect of succeeding |
| | no real prospect of success |
| **fanciful not real** | fanciful not real |
| | realistic as opposed to a fanciful |
| | real as opposed to a fanciful |
| | real and not merely fanciful |
| | more than fanciful |
| **mini trial** | mini trial |
| | mini-trial |
| | must not conduct a mini-trial |

The approach was chosen for its efficacy in not only quantifying the frequency of terms across individual cases but also in tracing patterns of term co-occurrence. From a legal perspective, this is insightful because the co-occurrence of legally significant terms can indicate stronger relevance for summary judgment proceedings. For instance, a decision containing both terms "no real prospect" and "part 24" has a higher probability of being a summary judgment than a case that only contains more commonly used legal terminology such as references to the "civil procedure rules". Tracking combinations and relationships between salient terms provides a richer lens into case content and context beyond isolated keyword usage. Prior to the search, both the keywords and the cases were converted to lowercase to avoid any mismatch.

To delve deeper, we conducted an analysis to determine the number of cases where any of the terms appeared in isolation, exclusively to all other terms (Table 3). This helps in discerning the frequency and context of these terms within the entire set of cases.

In addition, we plot several Venn Diagrams to visualise the co-occurrence and non-co-occurrence of the most frequently found search terms in the dataset (Fig. 2, Fig. 3 and Fig. 4).

Leveraging the gathered data on co-occurrences of search terms, we constructed a detailed matrix of logic to the search terms that extends beyond the basic use of the keywords. For instance, cases that exclusively mentioned "CPR", "no real prospect" or "civil procedure rules" without other summary judgment-related keywords were categorised as not summary judgments. This distinction is informed by the legal expert understanding that genuine summary judgments typically involve a combination of specific terms rather than the isolated use of these common terms. For example, some cases were identified where the legal test used in summary judgment proceedings is also used for other procedural matters. This meant that the court would refer to summary judgment and use all the relevant keywords, but not be making a determination on a true summary judgment case. For instance, one case identified contained the following text:

> "The test for whether permission should be given or an amendment to a claim form under CPR 17.3 is summarised at paragraph 17.3.6 of the White Book as being the same as under Part 24 (Summary Judgment) and is whether there is a real rather than a fanciful prospect of the claim succeeding."[5]

Therefore, the matrix of search terms was updated to exclude certain types of cases that would include terms like these but not relate to summary judgment based on specific criteria:

- Any mention of various forms of "application to amend" forms or permissions related to claim forms or defence documents under CPR 17.3;

- References to "application to serve" documents outside the jurisdiction under CPR rr. 6.36, 6.37 and 6.38; and

- Situations where a judgment was being challenged or set aside, expressed through terms like "set aside a default judgment", "set aside or vary a judgment" or "set aside a judgment entered in default" in conjunction with references to "CPR 13" or "CPR 13.3".

This analysis served as a supplement to our previous frequency analysis of key terms, providing a broader context for understanding the significance and distribution of these terms across different decisions in the corpus. The final comprehensive search logic applied is detailed in Appendix 1.

---

[5] *Roberts & Anor v Fresco* [2017] EWHC 283 (Ch) (17 February 2017) [52].

Applying this refined search matrix to the subset of cases containing the term "summary judgment" (and similar) resulted in two datasets: SJ Matrix Matches (2,408 cases) and Non-SJ Matrix Matches (3,545 cases) for a total of 5,953 cases. To ascertain the content of these two datasets, we performed manual reviews on each of them.

### 4.1.4    Manual checks on the search matrix results

To confirm the accuracy of our datasets, a legal expert conducted manual checks on the results of applying the search matrix. We calculated the sample size needed to achieve a statistically representative sample of SJ Matrix Matches (2,408 cases), and Non-SJ Matrix Matches (3,545 cases). We apply a finite population correction (**FPC**) to adjust our sample size. The sample size for a 95% confidence level with a 5% margin of error for the 2,408 cases is 332 cases and for the 3,545 cases is 347 cases. These cases were randomly selected from their respective datasets using Python's random number generation function.

### 4.2    Using a Large Language Model

For our comparative method, we use Claude 2 to identify if a case is a summary judgment or not. To date, we are unaware of any use of a LLM, including Claude 2, for identifying subsets of legal cases.

Claude 2 is an advanced language model developed by Anthropic, designed to embody the principles of being "helpful, harmless and honest" (Anthropic 2023a). It operates on a neural network that has been trained on large datasets comprising natural language, enabling it to comprehend and respond to open-ended prompts with reasoning and common sense. The training data is carefully filtered to minimise biases and harmful content. The capabilities of Claude 2 are powered by a technique called deliberation networks. This method involves breaking down prompts into discrete stages: comprehension, reasoning and response generation. Each step is handled by specialised modules that deliberate over the optimal understanding or reply.

The principal reason for selecting Claude 2 is its ability to handle document sizes of up to 100,000 tokens, corresponding to around 75,000 words (Anthropic 2023a). A token is a basic unit of text and punctuation. Beyond this, Claude 2 excels in language comprehension and contextual understanding (Liu et al. 2023). It can parse questions and scenarios, capturing the intricacies of language (Chang et al. 2023). It also encompasses structured knowledge of a wide array of concepts, encompassing various typologies and classification frameworks. The ability to differentiate between Type I and Type II scenarios (i.e., summary judgment and non-summary judgment) may necessitate connecting disparate pieces of information and deducing their consequences, a task for which Claude's deliberative capabilities may be particularly suited. Furthermore, Claude 2 can elucidate its reasoning process in detail. Claude's explanatory capacity, combined with its strengths in contextual analysis and multi-step logical reasoning, positions Claude 2 as a well-suited contender for judgment-based tasks (Anthropic 2023a). For such comprehension, Claude 2 uses transformers—a form of neural network architecture that analyses word order and context (Elhage et al. 2021). The model description indicates that its responses employ algorithms to ensure variety, precision and pertinence. Nonetheless, Claude 2, like any AI, is subject to intrinsic constraints and might sometimes provide responses that are incorrect or lack meaning.

### 4.2.1 Prompt to identify summary judgments

To accurately identify cases involving summary judgment proceedings, we created a specific prompt for Claude 2. The foundation of this prompt is the keyword framework derived from the exploratory analysis of the dataset. This framework encapsulates the key legal terminologies and concepts pertinent to summary judgment cases.

The prompt, as detailed in Appendix 2, was crafted to guide Claude 2 in analysing legal judgments and determining if they involved summary judgment proceedings. The prompt was structured to encompass the following critical aspects of summary judgments. First, the nature of the application by ascertaining whether the case involved an application for summary judgment or an appeal from a decision granting or refusing summary judgment. Second, the scope of summary judgment by including instructions to identify if summary judgment was a central issue in a case or constituted an alternative application to a strike-out. Third, the assessment of legal merits by prompting Claude 2 to evaluate if the case involved a determination of "no real prospect of succeeding" on a claim or defence, and if there was "no compelling reason" for the case to proceed to a full trial under the CPR or other tribunal procedural rules.

An initial draft of the prompt was largely similar but did not contain examples for clarification or the prompt structure. Following the publication of prompting guidance from Anthropic (2023b, c), this prompt was optimised for the challenging task of identifying summary judgment cases. Adhering to Anthropic's structuring recommendations, XML tags <case_text> were strategically used to segregate instructional text from the legal case text for analysis. This structure ensured that the AI focused on the relevant text segments, enhancing its processing efficiency. In line with Anthropic's advice, clear examples were incorporated within <example></example> tags. These examples served as practical guides, significantly improving the model's ability to identify and reason about summary judgment cases accurately. The prompt was crafted to pre-empt and exclude complex or atypical cases that could potentially lead to erroneous classifications. Using this prompt, Claude 2 was able to successfully detect if a case that we randomly selected was a summary judgment or not.

The procedure for uploading cases was straightforward. We used the Claude 2 Chat function, which allows a user to attach a file and enter a prompt for analysis (Anthropic 2023a). To integrate the use of Claude 2 into our software, we opted to use the application programming interface (**API**). The prompt was then run on 15 cases through the Claude API. The API was markedly faster, with responses being almost immediate compared to the several seconds it took in the Chat function. However, a manual check of these results showed that 7 out of 15 cases were incorrectly labelled by the Claude API. As a consequence, we switched back to the Claude 2 Chat console, which provided the correct responses for the same cases. We repeated tests with the API only to find that the same incorrect results were being generated. Hence, we maintained the use of the Chat console for the duration of this research. For future work, further investigation into the disparity between responses generated by the API and Chat is warranted.

### 4.2.2 Performing manual checks on the results of Claude 2

We then ran the prompt over all cases in the RegEx SJ dataset. Once Claude 2 had concluded its analysis of all 5,953 cases, we divided the results into two datasets, one where Claude 2 identifies a case as a summary judgment (3,102 cases) and one where Claude 2 identifies it as not a summary judgment (2,811 cases). Claude 2 was unable to identify 40 cases as they were beyond the imposed word limit. They ranged from 70,961 words to 239,178 words and, therefore, exceeded the limit of 100,000 tokens (including the prompt tokens). Given that it was not possible to use these cases with the current capacity of Claude 2, these cases were omitted.

Manual checks of both datasets were conducted. To ensure a statistically significant sample, we calculated the necessary sample sizes by applying an FPC. The sample size for a 95% confidence level with a 5% margin of error for the 3,102 cases was 342 cases, and the sample size for the 2,811 cases was 339 cases. Samples were randomly drawn from their respective datasets utilising Python's random number generator.

## 5    Results

### 5.1    Keyword search logic based on RegEx

In this section, we focus on identifying and analysing the distribution of summary judgment cases using the RegEx approach. This first approach involved searching the CLC using the summary judgment RegEx pattern. The search yielded 5,953 cases. Table 2 below provides the number of counts for each keyword from the above Table 1 that appears in these cases. If a keyword appeared multiple times within one case, then it is counted multiple times.

*Table 2: Number of times a keyword is identified in the RegEx resulting dataset*

| Keyword | Count |
|---|---|
| summary judgment | 27061 |
| no real prospect | 6012 |
| real prospect of success | 5576 |
| no real prospect of success | 2843 |
| cpr 24 | 1616 |
| real prospect of succeeding | 1533 |
| no real prospect of succeeding | 1330 |
| cpr part 24 | 1244 |
| mini-trial | 1213 |
| r 24.2 | 1115 |
| cpr 24.2 | 1009 |
| realistic prospect of success | 914 |
| summary judgement | 531 |
| part 24 application | 456 |
| ewhc 339 (ch) | 431 |
| rule 24.2 | 389 |
| compelling reason why the case or issue should be disposed of at a trial | 278 |
| mini trial | 256 |
| easyair ltd v opal telecom | 238 |
| more than fanciful | 156 |
| part 24 of the civil procedure rules | 90 |
| realistic prospect of succeeding | 74 |
| r. 24.2 | 56 |
| real as opposed to a fanciful | 56 |

| easyair v opal | 38 |
| realistic as opposed to a fanciful | 38 |
| must not conduct a mini-trial | 36 |
| part 24 judgment | 31 |
| civil procedure rules part 24 | 7 |
| easyair ltd. (t.a openair) v. opal telecom ltd [2009] ewhc 339 (ch) | 3 |
| real and not merely fanciful | 2 |
| fanciful not real | 1 |
| compelling reason to try the case or issue | 0 |
| part 24 judgement | 0 |

While these figures give insight into the use of these keywords in the data, a more characterising approach was applied using co-occurrence counts. Co-occurrence counts refer to how often the keywords appear together within the set of cases. This can be particularly insightful in understanding the relationships and common contexts between the different keywords. A co-occurrence happens when two keywords are found in the same document. For instance, if the keywords "summary judgment" and "cpr part 24" appear in the same case file, this is counted as a co-occurrence. If keywords co-occur multiple times within one case file, then this co-occurrence is only counted once. Counting keyword co-occurrences only once per case file shifts the focus from the sheer quantity of keyword mentions to the relevance of documents based on the diversity of keyword presence. Using this data, we constructed a co-occurrence matrix (Fig. 1), which details how frequently different keyword search terms, as defined in Table 1, appear together in this dataset. In Fig. 1 both rows and columns represent the keyword search terms. Each cell in the matrix represents the relationship between the keyword search terms. For instance, "summary judgment" and "cpr part 24" intersect to give 815 cases in which both these keywords co-occur at least once. The cell values along the diagonal, where the row and column represent the same keyword, display how many cases a particular keyword occurs at least once. For example, where "summary judgment" is on both the row and column, we find 5,819 cases.

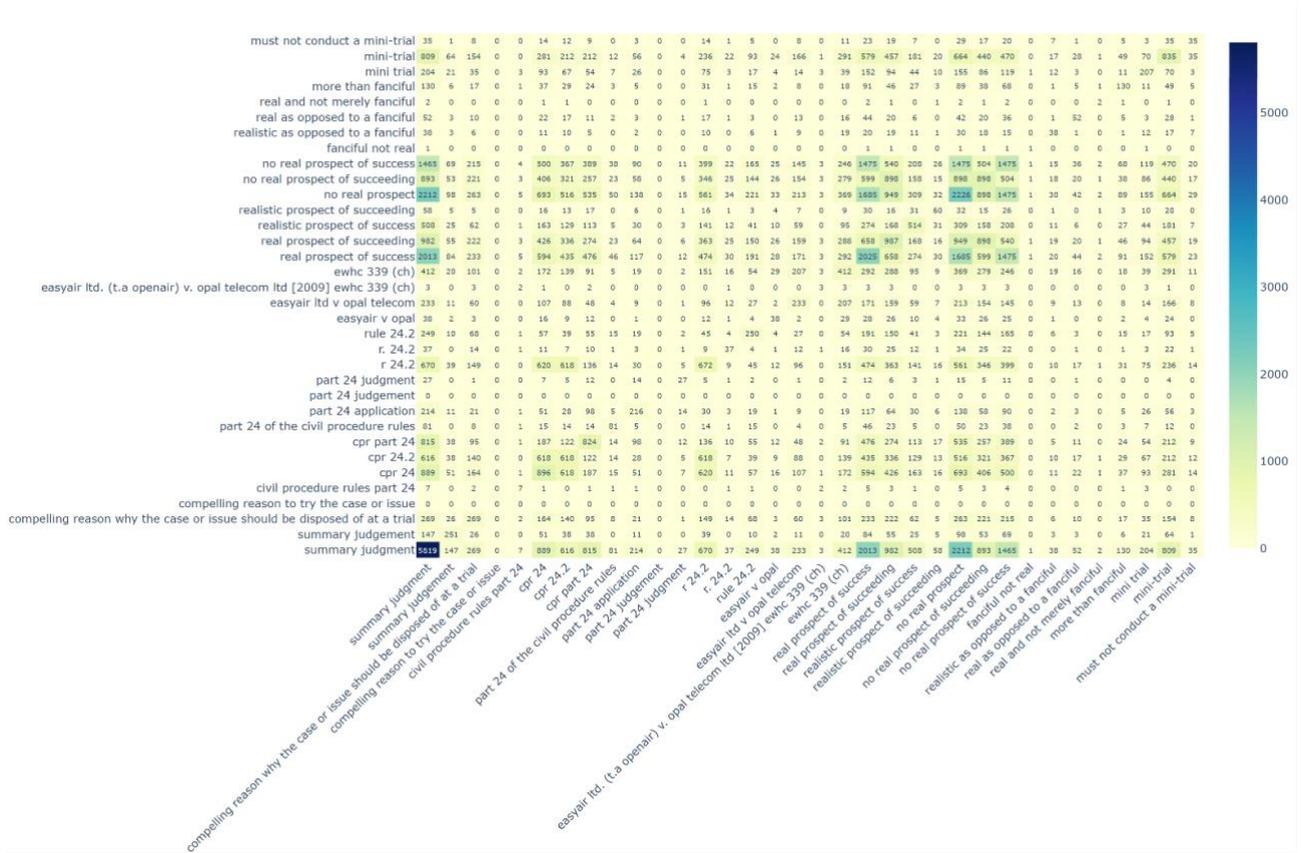

*Fig. 1: Co-occurrence matrix for all keywords*

As mentioned earlier, searching for the keyword "summary judgment" does not necessarily result in actual summary judgment decisions. To test this, we performed a unique count of the number of times a keyword appeared in isolation within a case (Table 3). This table highlights that the majority of terms co-occur, with only two exceptions. As can be seen in Table 3, there are 2,511 cases that contain "summary judgment" and no other keyword. According to legal expert knowledge, these cases are very likely not summary judgment cases. Instead, summary judgment cases will contain additional keywords derived from the legal test that must be applied by the court to make such a decision. This finding underscores that the term "summary judgment" can be mentioned in a case without referencing other elements typically associated with a summary judgment case. We also note that all other keywords, for example, "no real prospect of success", always co-occurred with other terms.

*Table 3: Number of times a keyword occurs in isolation*

| Term | Count |
|---|---|
| summary judgement | 4 |
| summary judgment | 2511 |

With respect to the clusters of co-occurring keywords, we find that some terms occur more frequently with others and, at times, never with others. For example, in Fig. 2, a Venn diagram illustrates the co-occurrence of the keyword categories "summary judgment", "no real prospect" and "mini-trial". The Venn diagram has three circles, each representing one of the keywords under consideration. The overlapping areas between circles represent the number of cases in which keywords co-occur. The central intersection of all three circles shows cases where all keywords are found together. As in Fig. 1, multiple co-occurrences within one case file are only counted once. The non-overlapping sections of each circle represent decisions where that keyword appears without the others. We also show, in Fig. 3, the Venn diagram for "summary judgment", "no real prospect" and "cpr 24". Fig. 4 shows the Venn diagram for "summary judgment", "cpr 24" and "r 24.2".

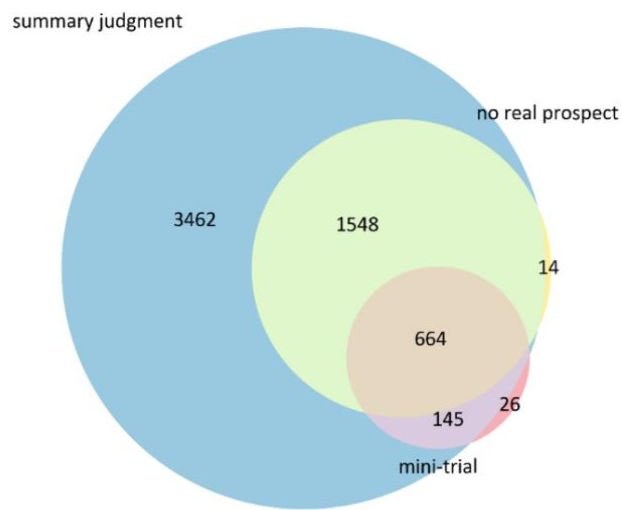

*Fig. 2: Venn diagram of co-occurrence of keywords "summary judgment", "no real prospect" and "mini-trial".*

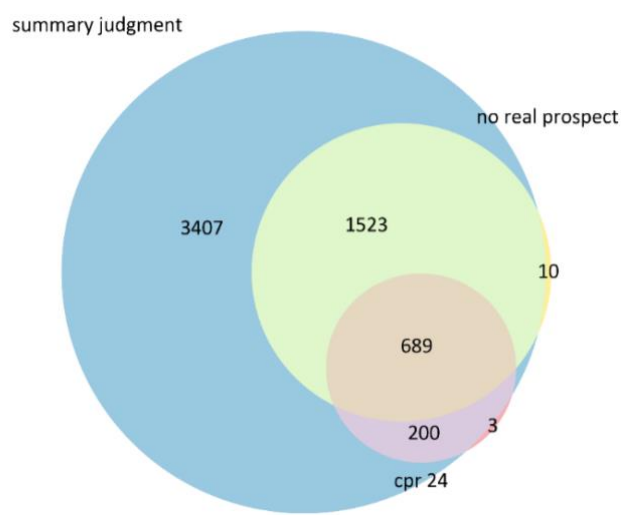

*Fig. 3: Venn diagram of co-occurrence of keywords "summary judgment", "no real prospect" and "cpr 24"*

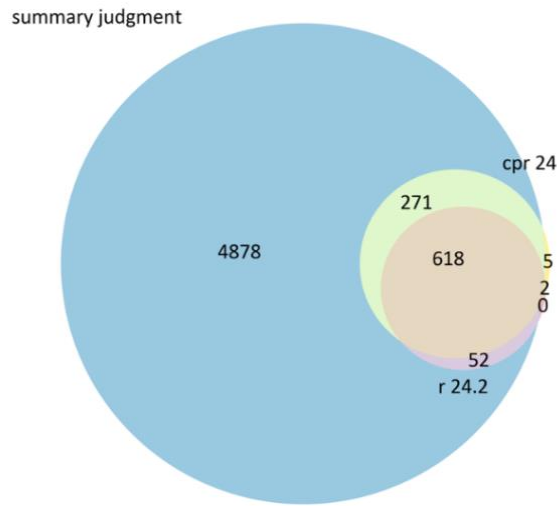

*Fig. 4: Venn diagram of co-occurrence of keywords "summary judgment", "cpr 24" and "r 24.2"*

## 5.2    Distribution of SJ cases based on RegEx

Applying the keyword search logic in Appendix 1 to the 5,953 cases that were identified using the summary judgment RegEx searches, we obtained 2,408 cases. We call this the Keyword SJ Dataset (**KSJD**). Concomitantly, this search also yielded 3,545 cases that did not match the keyword search. We named this dataset the Keyword non-SJ Dataset (**KNSJD**).

A statistically representative sample from both datasets was taken and checked by a legal expert (Table 4). The manual checks of the keyword datasets found that of the 332 summary judgment cases in the sample dataset, 64.8% were correctly identified. Of the 347 non-summary judgment cases, 89.6% were correctly identified. We translated this into a confusion matrix, as given in Table 5. The confusion matrix provides an F1 score. It is a statistical measure used to evaluate the accuracy of a test. It considers both the precision (how many identified items are relevant) and the recall (how many relevant items are identified). Hence, it offers a better appraisal than just measuring correct instances, providing a more holistic view of a test's effectiveness, especially in uneven datasets.

*Table 4: Manual checks of the RegEx SJ dataset*

| RegEx SJ dataset | Incorrect | Correct | Total Reviewed | Correct % |
|---|---|---|---|---|
| **SJ Matrix Matches** | 117 | 215 | 332 | 64.8 |
| **Non-SJ Matrix Matches** | 36 | 311 | 347 | 89.6 |

*Table 5: Confusion matrix based on manual checks of the RegEx SJ dataset*

| RegEx SJ dataset | Predicted SJ | Predicted Non-SJ |
|---|---|---|
| **Actual SJ Cases** | 215 | 36 |
| **Actual Non-SJ Cases** | 117 | 311 |

The confusion matrix gives an F1 accuracy score of 0.74 out of 1.00. Adjusting for the imbalance using a weighted F1, we get an F1 score of 0.78 out of 1.00. This concluded the categorisation based on expert knowledge to identify summary judgment cases.

*5.3    Distribution of SJ cases based on Claude*

Given the sub-optimal accuracy in the RegEx SJ dataset, we used Claude 2 to categorise cases with the aim of achieving a higher accuracy. The results of the Claude 2 classification based on the prompt in Appendix 2 yielded 3,102 cases that were identified as summary judgment (Claude SJ Dataset, **CSJD**). 2,811 cases were categorised as not summary judgment (Claude Non-SJ Dataset, **CNSJD**).

We created a statistically representative sample for both datasets and conducted manual checks (Table 6). We find that of the 342 cases, 90.40% were correctly classified as summary judgment, while of the 339 cases, 98.20% were correctly identified as not summary judgment cases. On this basis, we present the confusion matrix below (Table 7).

*Table 6: Manual checks of the Claude 2 dataset*

| Dataset | Incorrect | Correct | Total Reviewed | Correct % |
|---|---|---|---|---|
| **Claude identified SJ cases** | 33 | 309 | 342 | 90.4% |
| **Claude identified non-SJ cases** | 6 | 333 | 339 | 98.2% |

*Table 7: Confusion matrix based on manual checks of Claude 2 dataset*

| | **Predicted SJ** | **Predicted Non-SJ** |
|---|---|---|
| **Actual SJ Cases** | 309 | 6 |
| **Actual non-SJ Cases** | 33 | 333 |

This gives an F1 score of 0.94 out of 1.00 for Claude. Adjusting for class imbalance using weights, we get an F1 score of 0.94 out of 1.00. The significance of this result is covered in the discussion.

*5.4    Distribution of SJ cases per court*

As we are interested in analysing genuine summary judgment cases, we used the more accurate Claude 2 dataset. We used the name of the court for each case, the date of hearing and the court tier list in Appendix 3 for our analysis. Some of the courts listed in the CLC produced no matching cases and are, therefore, not contained in the list in Appendix 3. The following visualisations aid in understanding the distribution and prevalence of summary judgment cases across different court levels and over time.

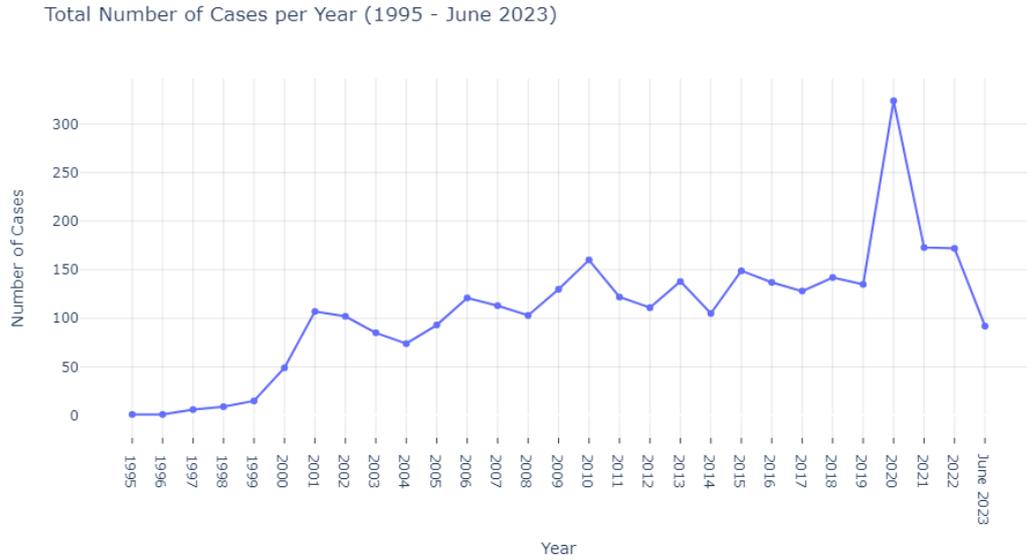

*Fig. 5: Total number of summary judgment cases per year (1995 to June 2023)*

First, we plot the aggregate number of summary judgment cases per year. Fig. 5 shows the increase of summary judgment cases after the implementation of the new CPR in 1997 and then a considerable peak in summary judgment cases during 2020, which might be the result of the need to resolve proceedings quickly during the COVID-19 pandemic. COVID-19 resulted in an increase in cases and a backlog within the court system. In particular, the number of cases where one or both parties were unrepresented increased significantly (Select Committee on the Constitution 2021). It follows that parties may have tried to resolve cases without a full trial in these circumstances.

We then identify the distribution of these summary judgment cases across the different courts (Table 8).

*Table 8: Number of summary judgment cases by court (1995 to June 2023)*

| Court | Number of Cases |
|---|---|
| England and Wales High Court (Chancery Division) | 769 |
| England and Wales Court of Appeal (Civil Division) | 640 |
| England and Wales High Court (Queen's Bench Division) | 543 |
| England and Wales High Court (Commercial Court) | 505 |
| England and Wales High Court (Technology and Construction Court) | 402 |
| England and Wales High Court (Patents Court) | 44 |
| England and Wales High Court (King's Bench Division) | 31 |
| Intellectual Property Enterprise Court | 24 |
| England and Wales High Court (Administrative Court) | 18 |
| The Judicial Committee of the Privy Council | 17 |
| First-tier Tribunal (Tax) | 16 |
| United Kingdom Competition Appeal Tribunal | 14 |
| United Kingdom Supreme Court | 12 |
| England and Wales Patents County Court | 10 |
| English and Welsh Courts - Miscellaneous | 10 |
| United Kingdom House of Lords | 10 |
| England and Wales Land Registry Adjudicator | 9 |
| England and Wales High Court (Admiralty Division) | 8 |
| England and Wales High Court (Family Division) | 4 |
| United Kingdom Employment Appeal Tribunal | 3 |
| First-tier Tribunal (General Regulatory Chamber) | 3 |
| England and Wales Leasehold Valuation Tribunal | 2 |
| United Kingdom Upper Tribunal (Tax and Chancery Chamber) | 1 |
| United Kingdom VAT & Duties Tribunals (Excise) | 1 |
| England and Wales High Court (Business and Property Courts) | 1 |
| First-tier Tribunal (Property Chamber) | 1 |
| England and Wales High Court (Mercantile Court) | 1 |
| England and Wales Court of Protection | 1 |
| England and Wales Court of Appeal (Criminal Division) | 1 |
| United Kingdom Employment Tribunal | 1 |

Based on this data, we then plot the number of summary judgment cases for selected courts in a line graph (Fig. 6) and a bar graph (Fig. 7). Additionally, we segment the list of UK courts into court tiers defined in Appendix 3 and plot the number of summary judgment cases by court tier (Fig. 8).

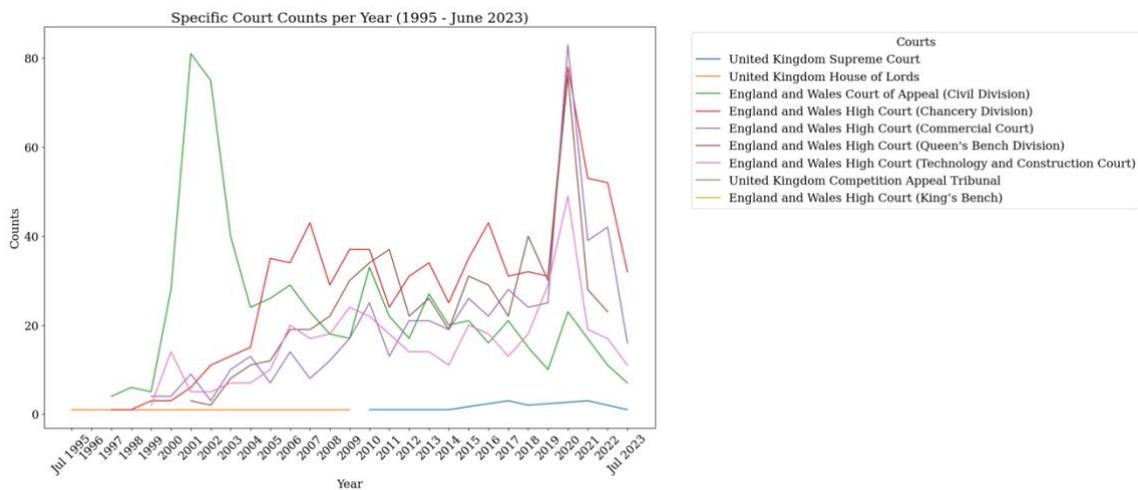

*Fig. 6: Line plot of the number of summary judgment cases by court*

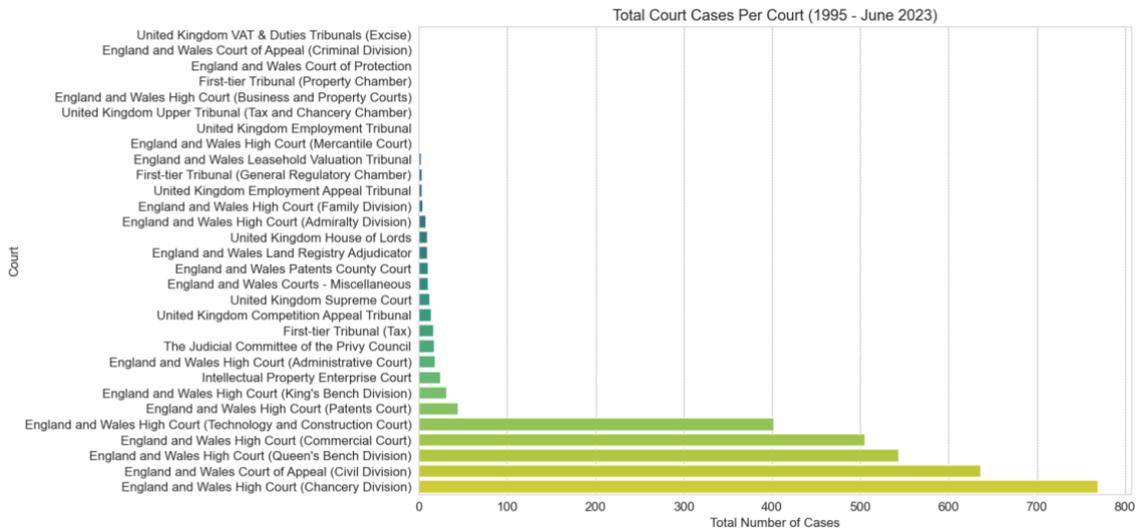

*Fig. 7: Bar graph of the number of summary judgment cases by court*

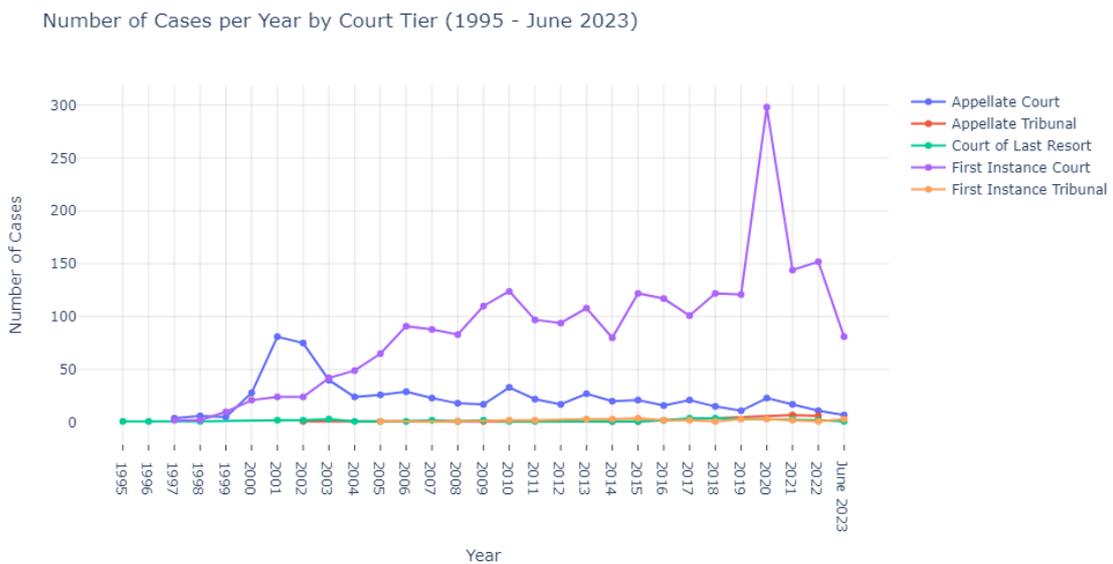

*Fig. 8: Number of summary judgment cases per year by court tier (1995-June 2023)*

We visually present the relationship between two key variables – the number of cases and the year – through a regression plot (Fig. 9). This plot includes a linear regression line, which is used to illustrate the trend in the data over time. This allows us to consider whether there is a noticeable increase or decrease in the number of cases each year.

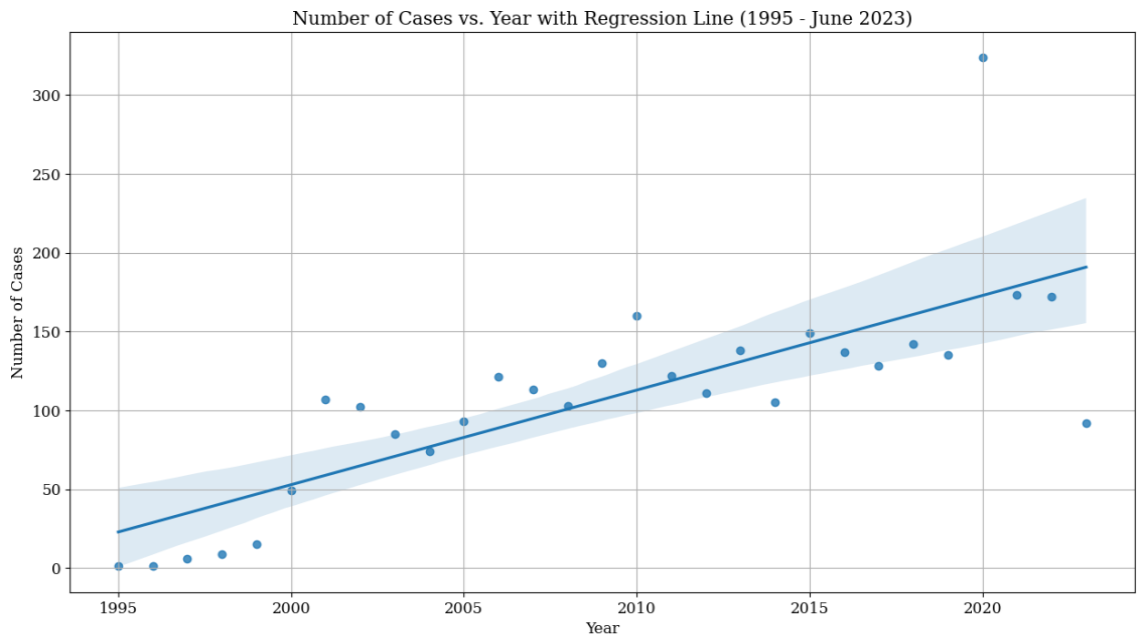

*Fig. 9. Regression plot for the number of cases and year*

The plot is found to have an r-squared value of 0.60. Fig. 9 indicates the extent to which the changes in the year can explain variations in the number of cases. An r-squared value of 0.60 means that 60% of the fluctuation in case numbers can be accounted for by the year. This is a significant percentage, suggesting a strong relationship between the time period and the number of cases. Additionally, the slope of the regression line is 6.00. This slope value quantifies the rate of change in case numbers over time. In practical terms, it means that each year, there is an average increase of 6 cases. The p-value is found to be 0.00, determining the result to be statistically significant. A p-value this low indicates that the probability of this trend occurring by chance is almost zero, thereby lending strong statistical support to the conclusion that there is a real, positive relationship between the year and the number of cases.

*5.5    Word count of summary judgment cases*

We then turn to analyse the word count of summary judgment cases. To start, we plot the average word count of summary judgment cases by year (Fig. 10). We then decompose the plot into its respective tiers by year (Fig. 11). Fig. 11 demonstrates that key information can be missed when aggregating by year across all courts.

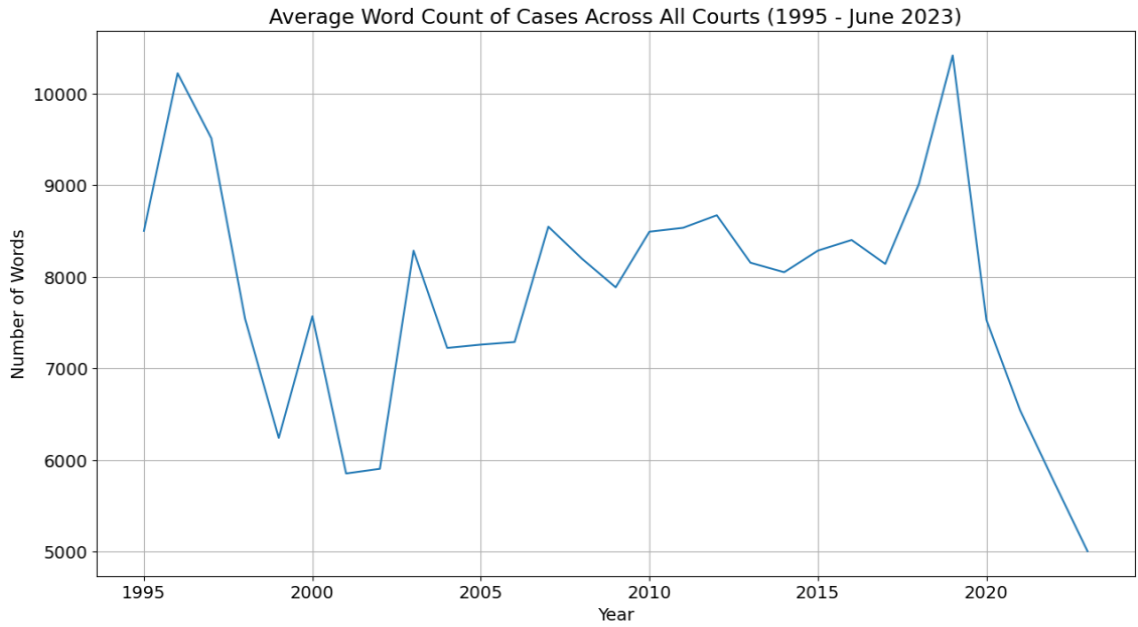

*Fig. 10: Average number of words in summary judgment cases by year (1995-June 2023)*

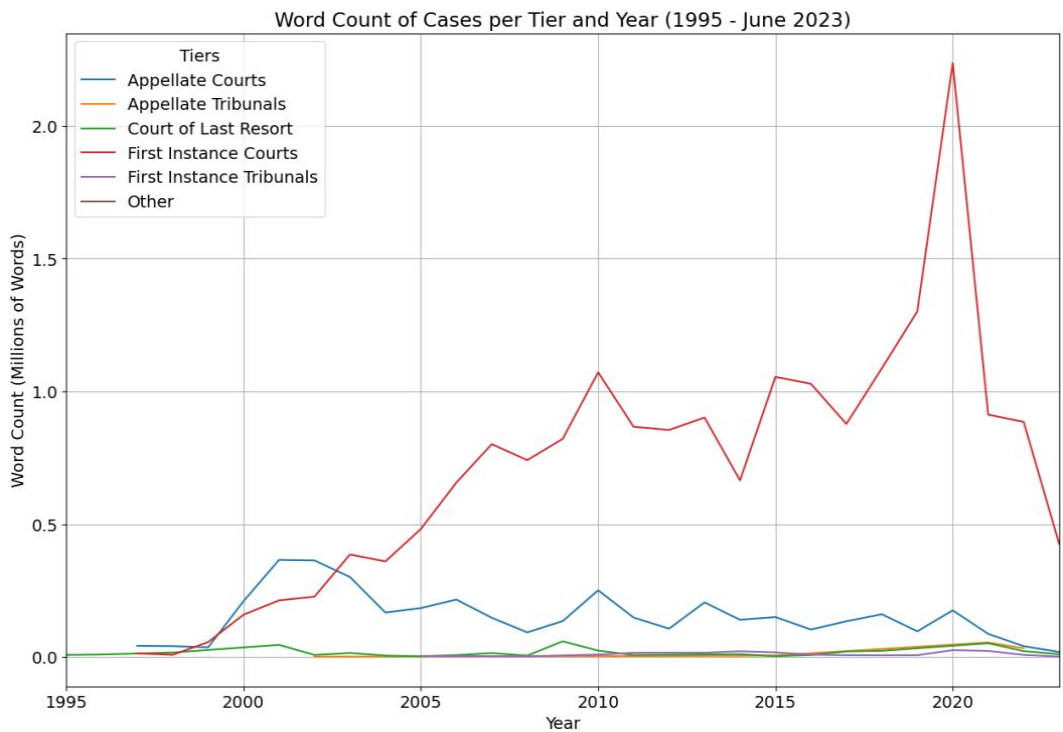

*Fig. 11: Number of words in summary judgment cases by year and by court tier (1995-June 2023)*

To further analyse our data, we then plot the word count for each summary judgment case per year and per court (Fig. 12). The lighter the colour, the higher word count.

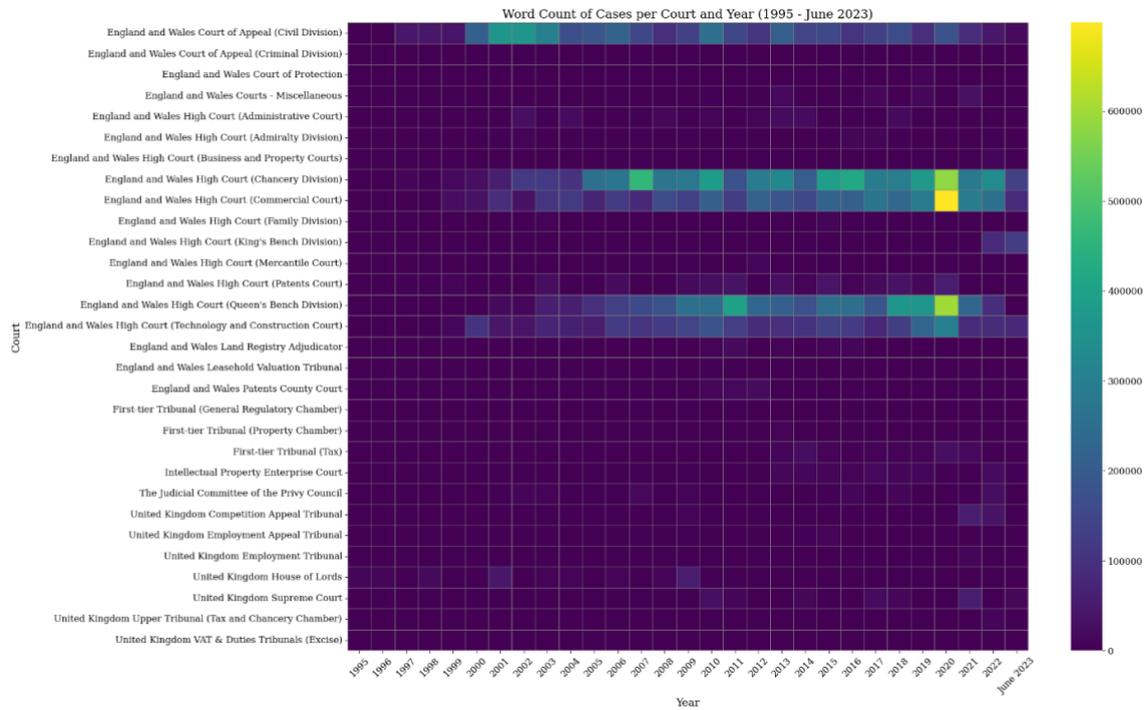

*Fig. 12: Number of words in summary judgment cases by year and by court*

While this seems to indicate an increase in the size of the cases within first instance courts, the average word count over the first instance courts and the average word count for all courts demonstrates less of an increase (Fig. 10, Fig. 11). Across all courts the word counts vary significantly, with a mean number of approximately 96,120 words. The standard deviation is relatively high at 121,843, indicating a wide dispersion in word counts across different entries. The minimum word count in a year is 303, and the maximum is 698,191.

Finally, we perform a K-means clustering on the word count results (Ikotun et al. 2023). Similar data points are clustered together in groups. In our case, cases with similar word counts are clustered together. Splitting the counts into two clusters, we find that Cluster 1 has cases with documents containing between 303 and 161,604 words, and that the majority of the cases (77.5%) fall into this group. The remaining cases fall into Cluster 2 which contains cases between 167,969 and 698,191 words representing 22.5% of cases.

It is also possible to break up the word count results into smaller clusters. Fig. 13 below shows a plot for 10 clusters. Notably, 49% of the cases appear within the first (blue) cluster. These represent cases with word counts between 303 and 27,496 words. The remaining cases fall within the remaining 9 clusters stratified as shown in Fig. 13.

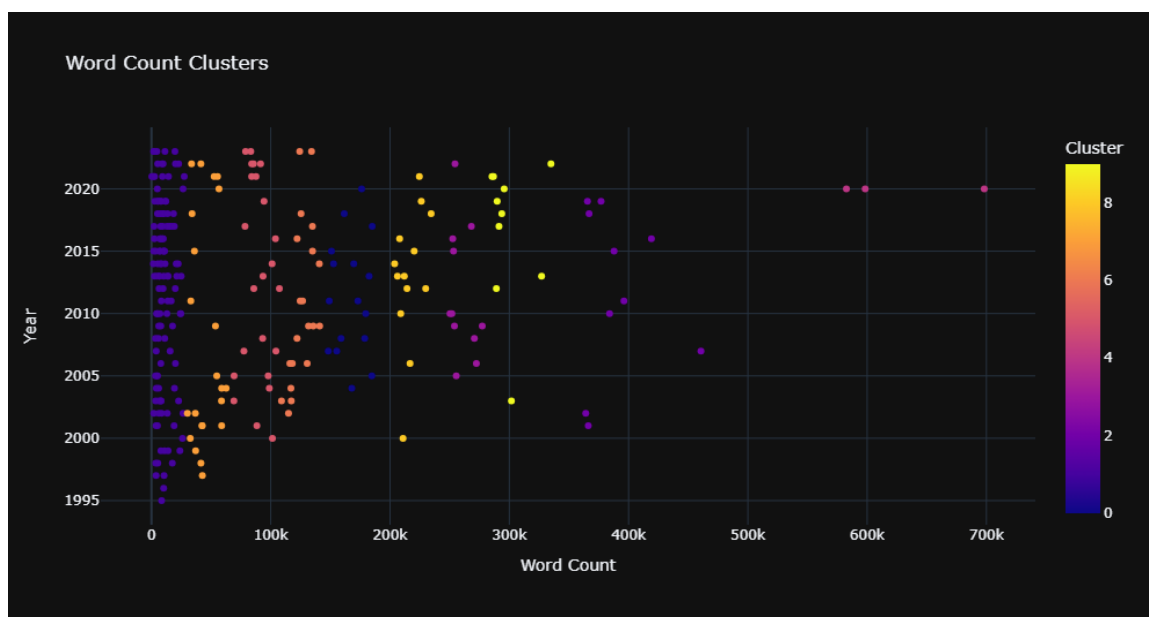

*Fig. 13: A scatter plot of the distribution of cases categorised into 10 clusters based on the word count of the documents. Each dot on the plot represents an individual case*

We next consider the quartiles. Quartiles are a statistical measure that divides a dataset into four equal parts, each representing a quarter of the dataset. They are used to understand the spread and distribution of data, especially in terms of its central tendency and variability. For our case, the 25th percentile (below which 25% of the data falls) is at 8,472 words, the median (50th percentile) is at 33,588 words (the median), and the 75th percentile is at 148,199 words. Of note, the average length of a court case in the whole of the CLC dataset is 5,824 words.

## 6      Discussion

### 6.1     Analysis of co-occurrences

The detailed examination of keyword co-occurrences in the summary judgments reveals interesting patterns that help guide our search approach, given the dynamic nature of legal terminology. Certain keywords appear repeatedly across the top co-occurrences. For example, the top co-occurring words are "summary judgment", "no real prospect" and "real prospect of success".

The CPR Rule 24.3(a) provides that as a part of its adjudication, the court may consider whether a "party has no real prospect of succeeding on the claim, defence or issue". We identified frequent occurrences of several variations on this test like "real prospect of succeeding", "real prospect of success", "no real prospect", "no real prospect of success" and "no real prospect of succeeding". This highlights the nuanced and often subtle differences in legal language that can significantly impact the identification of keywords. The variability of English legal language is a broader issue in automated legal research due to the complex and evolving lexicon within legal judgments.

The Venn diagrams illustrate how using different combinations of categories can bring about drastically different numbers of cases. In considering the Venn diagrams, we find that the three keywords in Fig. 2 co-occur 664 times. When we consider the two keywords "summary judgment" and "no real prospect", the co-occurrence increases to 1,548, i.e., a change of 884 cases. However, in only considering the keywords "no real prospect" and "mini-trial" the figure drops to 664 cases. The same real prospect of success test is applied by courts to consider other rules within the CPR, including applications to amend forms under CPR 17.3, applications to serve documents outside of jurisdictions under CPR rules 6.36, 6.37 and 6.38, and applications to set aside or vary a default judgment under CPR 13. Importantly, the reason for the large number of co-occurrences between "summary judgment" and "no real prospect" without the third keyword "mini-trial" that is unique to CPR 24, is that when applying for permission to appeal, the court will consider whether "the appeal would have a real prospect of success" (CPR 52.6). Therefore, some cases that mention "summary judgment" may be the basis of a permission to appeal application rather than a substantive summary judgment case. Whereas the third keyword "mini-trial" is usually more limited to true summary judgment proceedings. This variation is also seen when we consider the three keywords: "summary judgment", "no real prospect" and "cpr 24" (Fig. 3), which intersect 689 times. When we consider the intersection of "no real prospect" and "cpr 24" in Fig. 2, instead of "no real prospect" and "mini-trial" in Fig. 3, there is an increase by 25 cases. Finally, in Fig. 4 the keywords "summary judgment", "cpr 24" and "r 24.2" occur together in 618 cases and the overlap between "cpr 24" and "r 24.2" shows a large intersection, which is interesting since they are both references to codified law, i.e., the written procedural rules. The keywords "summary judgment" and "cpr 24", the broader reference, co-occur 889 times, compared to "summary judgment" and "r 24.2, the narrower reference, which co-occur 670 times. Such occurrences arise because we use more nuanced terms employed in summary judgment cases in each iteration of the Venn diagrams.

These figures indicate that the use of keywords to find summary judgment can be challenging. At the same time, the keyword analysis greatly aided the development of the search term logic and prompt in Appendix 1 and Appendix 2 respectively. These co-occurrences may also be valuable for predictive analysis in legal research or practice.

*6.2    Analysis of using domain knowledge and LLMs*

The use of logical operators to find information in legal databases is well established. However, a systematic benchmark of methods that utilise these operators with keywords to identify specific types of cases does not appear in the literature to the best of our knowledge. Likewise, benchmarking for the use of LLMs for this specific task is also not present in the literature. As a consequence, there is no clear comparator for our results. Our findings confirm a survey of 30 years of case law retrieval that concluded that the research comparing and improving search methods is lagging with no evaluation benchmark collections available (Locke and Zuccon 2022).

There are studies that have investigated the use of logic operators and natural language in case law information retrieval, with F1 scores reported. While these studies do not match our task, we mention them here for the closest available comparison. In a very comprehensive study on this topic, Turtle (1994) considers logic operators for searching private, commercial legal databases, Westlaw and LexisNexis. This research finds an F1 score of 0.309 when considering the first 20 results. Turtle (1994) performing a secondary study on a larger collection finds an F1 score of 0.32. The study was repeated using natural language queries. The results from the comprehensive study gave improved results, with an F1 score of 0.417 for a small collection and F1 of 0.397 for a larger collection. Dabney (1986) performs a logical operator search using Westlaw and obtains an F1 score of 0.227 and F1 of 0.159 for LexisNexis.

These results contrast with the higher scores found using both methods in this paper. Nevertheless, it remains that the LLM produced more precise and accurate results. Interestingly, both the RegEx- and the LLM-based approaches achieve higher accuracy in identifying non-SJ cases than in recognising SJ cases.

However, despite the very high accuracy of using an LLM, Claude 2 still did not identify 9.6% of summary judgment cases, as shown in Table 6. Our results and visualisations are to be taken with this limitation. Claude 2 was also over-inclusive by incorrectly classing 33 non-summary judgement cases as summary judgment cases, as given in Table 7. It did, however, do better as regards misclassing actual summary judgments as non-summary judgment cases (6 cases).

These findings reflect the fact that law is a field that relies heavily on the use of notoriously complex, domain-specific language (Ruhl 2008; Katz and Bommarito 2014; Friedrich 2021). The complexity of legal language and massive volumes of textual data have made it difficult to develop NLP systems that understand the nuances of legal tasks (Katz et al. 2020; Chalkidis et al. 2022). At this point in time, there will be a margin of error when using LLMs for complex tasks such as our task of understanding the subject matter and procedure of a legal case.

*6.3    Implications for legal research and practice*

Our methods have significant implications for legal research and practice. We find that the absence of standardised legal language in judgment reasoning, particularly in summary judgments, makes it difficult to identify an entire corpus of cases concerning one area of law or procedure.

For legal researchers, our findings offer a roadmap for navigating legal databases more efficiently. By recognising frequently co-occurring terms, researchers can refine search strategies, leading to more accurate and faster retrieval of relevant case law. This is particularly valuable in legal education and for new practitioners, providing them with insights into the lexicon commonly used in judicial decisions.

Furthermore, these patterns could inform the development of AI-based legal research tools. Such tools, leveraging the predictive power of keyword co-occurrences, can revolutionise legal research by offering more precise and contextually relevant search results. This aligns with the growing trend of digital transformation in the legal field, where technology is increasingly integrated to enhance the efficiency and accuracy of legal work.

Our results also underscore the superiority of LLMs over traditional NLP methods in tasks such as information extraction and text classification (Gartlehner et al. 2023), as demonstrated in the area of legal case analysis. Zhao et al. (2023) highlight the enhanced capabilities of LLMs in processing unstructured text and extracting structured data, with significant advancements over previous methodologies. Specifically, LLMs overcome key challenges in named entity recognition, relation extraction and semantic parsing (Perot et al. 2023). They also achieve strong performance at zero- and few-shot information extraction, even when working with texts from specialised domains such as medicine, without domain-specific fine-tuning (Agrawal et al. 2022).

The principal reasons why LLMs excel at these tasks are as follows (Minaee et al. 2024):

i.  Pretraining: LLMs are pre-trained on very large datasets using approaches like autoregressive language modelling. This process enables the model to learn to predict the next word in a sentence given the words that have come before it. This form of training helps the model understand language structure, grammar and context. This, in turn, facilitates a rich representation of linguistic knowledge.

ii. Larger model capacity: LLMs have typically use extremely large numbers of parameters which allow them to learn more complex language patterns and generalise effectively.

iii. Effective prompting: LLMs can employ in-context learning, whereby the models learn from data presented with contextual information. This allows the model to interpret the prompt and leverage its contextual knowledge.

iv. Emergent abilities: As the model scale increases, LLMs exhibit emergent generalisation skills. This occurs as the model begins to detect latent patterns in the data, allowing it to represent its learning at higher abstractions. Such higher abstraction allows for better generalisation.

One challenge with LLMs is that they hallucinate (Andriopoulos and Pouwelse 2023; Li 2023; Agrawal et al. 2023). Efforts are underway to limit or control this phenomenon in the legal domain. For example, ChatLaw integrates external knowledge bases to overcome model hallucinations during legal data retrieval (Cui et al. 2023). It demonstrates that using in-domain tuning data and grounding can mitigate risks such as false information.

## 6.4 Limitations

This study, while utilising the most comprehensive dataset of UK court cases currently available, acknowledges certain limitations. Notably, the dataset may not encompass all summary judgments delivered by the courts, as explained in Section 3. This potential gap in data coverage could influence the findings, particularly in the context of keyword co-occurrence and term frequency analysis. Such limitations highlight the need for continuous data collection and updates to ensure the dataset remains representative and comprehensive. The limitation of the dataset is that statistically significant sampling determined that 9.6% of the cases were non-summary judgment cases.

## 6.5 Further work

Future research directions could focus on refining the methodology to enhance accuracy and reliability. This includes exploring prompt engineering strategies to improve the F1 score, a measure of a test's accuracy. The application of LLMs to provide feedback for prompt refinement presents an innovative approach. Additionally, employing traditional classification methods with a manually curated dataset could verify the feasibility of achieving higher F1 scores and provide a comparative analysis with existing models like Claude 2. This approach would not only validate the current findings but also offer insights into the evolving landscape of legal data analysis.

# 7 Conclusion

This study has successfully demonstrated an innovative application of NLP and AI in legal research. By confronting the challenge of identifying and categorising summary judgment cases within an expansive corpus of UK court and tribunal decisions, we have provided a novel dual-methodological approach that combines the benefits of expert-derived keywords with the robust contextual use of an LLM.

Our findings indicate that both methods offer advantages that can be used complementarily to refine the search for subsets of legal cases. Results from the keywords can inform better prompts. In addition, the precision achieved by prompting Claude 2 is a considerable contribution of this paper. The keyword-based approach, grounded in expert legal understanding, ensures the retrieval of cases based on an established legal lexicon. At the same time, Claude 2 introduces an element of contextual discernment, allowing for a broader yet more precise classification that mirrors human-like reasoning. The ability to identify summary judgment cases through these methods addresses our primary research question.

Exploring our second research question, the distribution of SJ cases across UK courts reveals patterns that merit further investigation. While summary judgments were used in 30 UK courts, the majority were first instance courts. Five courts used the procedure the most: England and Wales High Court (Chancery Division), England and Wales Court of Appeal (Civil Division), England and Wales High Court (Queen's Bench Division), England and Wales High Court (Commercial Court) and England and Wales High Court (Technology and Construction Court). A linear regression determined that there is an increase of 6 cases per year. The number of cases has been rising steadily. From 2001 to 2009 the range of cases has been within 100 to 150 per year. This was disturbed by one spike in 2020 possibly due to COVID-19 and the ability of SJ to find a quick solution to a conflict. Our data shows that the median word count is 32,768, and that 48.8% of the summary judgement cases have word counts between 303 and 26,479 words.

Our contributions to the field are twofold: we establish a validated method for isolating SJ decisions from large case law corpora and provide an empirical foundation for subsequent studies to investigate trends in summary judgments. This work addresses concerns with respect to access to justice and the potential role of AI in democratising legal information.

The paper demonstrates the potential to unravel complex legal threads with increased efficiency and clarity by leveraging modern technological advancements in natural language processing and artificial intelligence. The methods in this paper are applicable to any information extraction pipeline in the specialised domain of law. The processes also demonstrate how AI can augment the capabilities of legal professionals and researchers.

**Appendix 1 – Summary Judgment Search Matrix**

We have organised a search matrix based on the summary judgment keywords in Table 1 above. Table 9 below demonstrates how these search terms are combined across these categories, with each column representing a unique combination of terms. The logic for each category is detailed below, indicating the specific combinations of terms used:

1. this is an application for summary judgment

2. SJ + compelling reason

    a. (summary judgment OR summary judgement) AND ("compelling reason why the case or issue should be disposed of at a trial" OR "compelling reason to try the case or issue")

3. SJ + CPR 24

    a. (summary judgment OR summary judgement) AND (civil procedure rules part 24 OR "cpr 24" OR "cpr 24.2" OR "cpr part 24" OR "part 24 of the civil procedure rules" OR "part 24 application" OR "part 24 judgement" OR "part 24 judgment" OR "r 24.2" OR "r. 24.2" OR "rule 24.2" OR "summary judgment application")

4. SJ + easyair v opal

    a. (summary judgment OR summary judgement) AND ("easyair v opal" OR "easyair ltd v opal telecom" OR "easyair ltd. (t.a openair) v. opal telecom ltd [2009] ewhc 339 (ch)" OR "ewhc 339 (ch)")

5. SJ + real prospect of success + CPR 24

    a. (summary judgment OR summary judgement) AND ("real prospect of success" OR "real prospect of succeeding" OR "realistic prospect of success" OR "realistic prospect of succeeding" OR "no real prospect" OR "no real prospect of succeeding" OR "no real prospect of success") AND (civil procedure rules part 24 OR "cpr 24" OR "cpr 24.2" OR "cpr part 24" OR "part 24 of the civil procedure rules" OR "part 24 application" OR "part 24 judgement" OR "part 24 judgment" OR "r 24.2" OR "r. 24.2" OR "rule 24.2" OR "summary judgment application")

6. SJ + real prospect of success + fanciful not real

    a. (summary judgment OR summary judgement) AND ("real prospect of success" OR "real prospect of succeeding" OR "realistic prospect of success" OR "realistic prospect of succeeding" OR "no real prospect" OR "no real prospect of succeeding" OR "no real prospect of success") AND ("fanciful not real" OR "realistic as opposed to a fanciful" OR "real as opposed to a fanciful" OR "real and not merely fanciful" OR "more than fanciful")

7. SJ + real prospect of success + mini trial

    a. (summary judgment OR summary judgement) AND ("real prospect of success" OR "real prospect of succeeding" OR "realistic prospect of success" OR "realistic prospect of succeeding" OR "no real prospect" OR "no real prospect of succeeding" OR "no real prospect of success") AND ("mini trial" OR "mini-trial" OR "must not conduct a mini-trial")

8. EXCLUDE:

    a. ("application to amend the claim form" OR "application to amend a claim form" OR "application to amend the Defence" OR "an amendment to a claim form under CPR 17.3" OR "application for permission to amend");

b. ("application to serve outside the jurisdiction" OR "application for permission to serve outside the jurisdiction" OR "merits of the relevant claim under CPR r.6.37(1)(b)" OR "under CPR rr. 6.36, 6.37 and 6.38");

c. ("set aside a default judgment" OR "set aside or vary a judgment" OR "set aside a judgment entered in default") AND ("CPR 13" OR "CPR 13.3)

*Table 9: Combination of search terms*

| | **Primary Term** | **Category** | **Combination Terms** | **Further Combinations** |
|---|---|---|---|---|
| 1. | this is an application for summary judgment | this is an application for summary judgment | this is an application for summary judgment | |
| 2. | **summary judgment** / summary judgement | **compelling reason** | compelling reason why the case or issue should be disposed of at a trial | |
| | | | compelling reason to try the case or issue | |
| 4. | **summary judgment** / summary judgement | **civil procedure rules part 24** | civil procedure rules part 24 | |
| | | | cpr 24 | |
| | | | cpr 24.2 | |
| | | | cpr part 24 | |
| | | | part 24 of the civil procedure rules | |
| | | | part 24 application | |
| | | | part 24 judgement | |
| | | | part 24 judgment | |
| | | | r 24.2 | |
| | | | r. 24.2 | |
| | | | rule 24.2 | |
| 4. | **summary judgment** / summary judgement | **easyair v opal** | easyair v opal | |
| | | | easyair ltd v opal telecom | |
| | | | easyair ltd. (t.a openair) v. opal telecom ltd [2009] ewhc 339 (ch) | |
| | | | ewhc 339 (ch) | |
| 5. | **summary judgment** / summary judgement | **real prospect of success** | real prospect of success | *Also in combination with* **civil procedure rules part 24** |
| | | | real prospect of succeeding | |
| | | | realistic prospect of success | |
| | | | realistic prospect of succeeding | |
| | | | no real prospect | |
| | | | no real prospect of succeeding | |
| | | | no real prospect of success | |
| 6. | **summary judgment** / summary judgement | **fanciful not real** | fanciful not real | *Also in combination with* **real prospect of success** |
| | | | realistic as opposed to a fanciful | |
| | | | real as opposed to a fanciful | |
| | | | real and not merely fanciful | |
| | | | more than fanciful | |
| 7. | **summary judgment** / summary judgement | **mini trial** | mini trial | *Also in combination with* **real prospect of success** |
| | | | mini-trial | |
| | | | must not conduct a mini-trial | |

**Appendix 2 – Claude 2 Prompt**

\n\nHuman: I need you to analyse this legal case judgment and identify if it involves summary judgment proceedings. Summary judgment is a legal process where a court can decide a case or issue without a full trial.

Relevant details to identify about summary judgment are:
- whether there is no real prospect of succeeding on or defending the claim and there is no compelling reason why the case or issue should be disposed of at trial under CPR or other tribunal procedural rules
- if there is a real prospect of succeeding, a real and not merely fanciful prospect of success, and whether there is a compelling reason to try the case or issue
- if a party brings an application requesting summary judgment or appeals a case for an issue in an earlier decision to grant or refuse summary judgment
- that summary judgment may be one issue in a case or about the whole case, it may also be an alternative application to a strike-out

Please analyse the full text contained within the <case_text> </case_text> tags. If the case involves an application for summary judgment or an appeal of a summary judgment decision, respond with:

<response> Yes, this is a summary judgment case. Reason: [insert 1-2 sentences explaining why you identified it as a summary judgment case] </response>

If it is NOT a summary judgment case, respond: <response> No, this is not a summary judgment case. </response>

Do NOT include the following as summary judgment cases:
- enforcement of adjudicator's award under the Construction Act
- applications to amend a claim form under CPR 17.3
- applications for permission to serve outside jurisdiction under CPR rr. 6.36, 6.37 and 6.38
- applications to set aside a default judgment under CPR 13

These types of cases may discuss similar legal tests but are not summary judgment cases. Focus only on identifying true summary judgment cases.

Here are two examples:

<example> <case_text> The plaintiff applied for summary judgment on the grounds that the defendant had no real prospect of establishing their defence. The court considered whether there was need for a full trial. </case_text> <response> Yes, this is a summary judgment case. Reason: The case includes an application for summary judgment and discusses 'no real prospect of success' and the need for the issue to proceed to full trial. </response>

<example> <case_text> The plaintiff disputes liability in a contractual dispute. The defendant applied to amend their defence claim form. </case_text> <response> No, this is not a summary judgment case. </response> <example>

\n\nAssistant:

**Appendix 3 – Court Tier List**

| Court Category | Court Name |
|---|---|
| Tier 1:<br>Court of Last Resort | United Kingdom Supreme Court |
| | United Kingdom House of Lords |
| | The Judicial Committee of the Privy Council |
| Tier 2:<br>Appellate Court | England and Wales Court of Appeal (Civil Division) |
| | England and Wales Court of Appeal (Criminal Division) |
| Tier 2:<br>Appellate Tribunal | United Kingdom Employment Appeal Tribunal |
| | United Kingdom Competition Appeal Tribunal |
| Tier 3:<br>First Instance Court | England and Wales Court of Protection |
| | England and Wales Courts - Miscellaneous |
| | England and Wales High Court (Administrative Court) |
| | England and Wales High Court (Admiralty Division) |
| | England and Wales High Court (Business and Property Courts) |
| | England and Wales High Court (Chancery Division) |
| | England and Wales High Court (Commercial Court) |
| | England and Wales High Court (Family Division) |
| | England and Wales High Court (King's Bench Division) |
| | England and Wales High Court (Mercantile Court) |
| | England and Wales High Court (Patents Court) |
| | England and Wales High Court (Queen's Bench Division) |
| | England and Wales High Court (Technology and Construction Court) |
| | England and Wales Patents County Court |
| Tier 3:<br>First Instance Tribunal | England and Wales Land Registry Adjudicator |
| | England and Wales Leasehold Valuation Tribunal |
| | First-tier Tribunal (General Regulatory Chamber) |
| | First-tier Tribunal (Property Chamber) |
| | First-tier Tribunal (Tax) |
| | United Kingdom Employment Tribunal |
| | United Kingdom Upper Tribunal (Tax and Chancery Chamber) |
| | United Kingdom VAT & Duties Tribunals (Excise) |

**Appendix 4 – Code Availability**

The anonymised code used for the paper is made available at: https://github.com/Anaffinis/LLM-vs-Lawyers